\documentclass[conference]{IEEEtran}
\IEEEoverridecommandlockouts
\usepackage{cite}
\usepackage{amsmath,amssymb,amsfonts}
\usepackage{algorithmic}
\usepackage{graphicx}
\usepackage{textcomp}
\usepackage{xcolor}

\usepackage{soul}
\usepackage{url}
\usepackage{tabularx}
\usepackage{subfig}
\usepackage{textcomp}
\usepackage{multirow}
\usepackage{booktabs}
\usepackage{changepage}
\usepackage{caption}
\DeclareMathOperator{\softmax}{softmax}

\DeclareMathOperator{\distance}{distance}
\DeclareMathOperator{\Cat}{Cat}
\DeclareMathOperator*{\argmax}{argmax}

\DeclareMathOperator*{\V}{V}

\def\BibTeX{{\rm B\kern-.05em{\sc i\kern-.025em b}\kern-.08em
    T\kern-.1667em\lower.7ex\hbox{E}\kern-.125emX}}

\makeatletter
\def\ps@IEEEtitlepagestyle{%
\def\@oddfoot{\mycopyrightnotice}%
\def\@evenfoot{}%
}
\def\mycopyrightnotice{%

}

\begin{document}

\title{Clustering Human Mobility with Multiple Spaces}

\author{\IEEEauthorblockN{Haoji Hu}
\IEEEauthorblockA{\textit{University of Minnesota, Twin Cities} \\
huxxx899@umn.edu}
\and
\IEEEauthorblockN{Haowen Lin}
\IEEEauthorblockA{\textit{University of Southern California}\\
haowenli@usc.edu}
\and
\IEEEauthorblockN{Yao-Yi Chiang}
\IEEEauthorblockA{\textit{University of Minnesota, Twin Cities}\\
yaoyi@umn.edu}
}

\maketitle

\begin{abstract}
Human mobility clustering is an important problem for understanding human mobility behaviors (e.g., work and school commutes). Existing methods typically contain two steps: choosing/learning a mobility representation and applying a clustering algorithm to the representation. However, these methods rely on strict visiting orders in trajectories and cannot take advantage of multiple types of mobility representations. This paper proposes a novel mobility clustering method for mobility behavior detection. First, the proposed method contains a permutation-equivalent operation to handle sub-trajectories that might have different visiting orders but similar impacts on mobility behaviors. Second, the proposed method utilizes a variational autoencoder architecture to simultaneously perform clustering in both latent and original spaces. Also, in order to handle the bias of a single latent space, our clustering assignment prediction considers multiple learned latent spaces at different epochs. This way, the proposed method produces
accurate results and can provide reliability estimates of each trajectory’s cluster assignment. The experiment shows that the proposed method outperformed state-of-the-art methods in mobility behavior detection from trajectories with better accuracy and more interpretability. 
\end{abstract}


\section{Introduction} 

Understanding individual human mobility patterns has been an actively-studied topic in the past  decade~\cite{gonzalez2008understanding,jiang2012clustering,su2021unveiling,zhou2021you}.  Clustering human mobility is an important task to group ``similar'' mobility behaviors and provides insight towards detected mobility behaviors patterns. As individual trajectories (e.g. raw individual GPS data) are usually used as the proximity for mobility, clustering individual trajectories to detect the mobility behaviors is widely studied~\cite{jiang2012clustering,yao2017trajectory,yao2018learning,yue2021vambc}. This direction is beneficial to many domains like policymaking, urban design, economics, and geo-spatial intelligence. 


Existing trajectory clustering methods follow a paradigm that first chooses a way to represent the trajectory and then applies a clustering algorithm on the trajectory representation to group similar trajectories~\cite{jiang2012clustering,yao2017trajectory,yao2018learning,yue2019detect,yue2021vambc,su2021unveiling,zhou2021you}. Although many advances have been achieved, there are still major challenges. 

First, even though there are different ways to represent a  trajectory, they fail to properly handle the diversity of multi-scale trajectories within the same mobility type. The core idea of the existing methods is either directly representing a trajectory as a temporal sequence of which the order between consecutive elements (e.g., recorded geocoordinates or stay points)  represents the transition between corresponding consecutive time steps in the trajectory~\cite{besse2016review,li2008mining} or learning a vector representation encoding all the elements and transitions from the temporal  sequence~\cite{yao2017trajectory,yue2021vambc,yue2019detect,zhou2021you}. The definition of an element varies across methods but the transition direction always represents the temporal order. The assumption of existing methods  is that \emph{the  same mobility behavior type  usually have  similar  element transitions}. As a transition  involves two  consecutive  elements and the direction between them, the  similarity is also based on both the  sequence elements and the transition direction between them. This idea has been  widely-used, but it ignores that some sub-trajectories of the same mobility behavior type may  have the same consecutive elements but with an opposite transition direction. For example, assuming  people in a school are asked  to get vaccines in the same hospital. A teacher starts the trip from the office (which is close to the bus station).  Then, the teacher takes a long walk to the parking ramps and drives to the hospital. And a student starts the trip from the classroom (which is close to the parking ramps).  Then, the student walks to the bus station and takes a bus to the hospital. Even though the teacher and the student have the same mobility goal (i.e., go to the hospital), their beginning sub-trajectories have the exact opposite element of transition.  


Second, it is unclear what are the best similarity measurement and space to calculate the similarity for mobility behavior clustering. Even though the goal is to group ``similar'' mobility behaviors, it is difficult to define what is the desired similarity measurement in an unsupervised setting. Existing methods simply use a predefined measurement (e.g., the Euclidean distance) as the proxy similarity and calculate the similarity in a single representation space (either the original data space or reduced/latent space)~\cite{xu2015comprehensive}. If the original space is suitable for clustering objective, the clustering algorithm can be directly applied. If the original space is not good enough (e.g., the original space suffers from the curse of dimensionality or the data in the original space distributes in a complex and clustering-unfriendly way), the data is first mapped into a reduced/latent space (e.g., using  principal component analysis (PCA) to linearly reduce the feature dimensions~\cite{xu2015comprehensive} or neural networks to nonlinearly transform~\cite{yang2017towards}  the data from the original space to a latent space~\cite{xie2016unsupervised}). Then, clustering with the predefined similarity is applied in the reduced/latent space. Usually, both options are tested empirically if it is not obvious to observe the limitation in the original space  (e.g., the original space has moderately large dimensions). Yet, it still is difficult to tell which representation space could provide better clustering results when we do not have the ground truth labels (i.e., unsupervised clustering). Also, there is no guarantee that choosing one representation could be enough for us to ignore the other representation without losing any useful information for the clustering membership assignment. For example,  PCA helps battle the curse of dimensionality, but it is an information-loss transformation.  It is unclear if we lose useful information. Even though nonlinear transformation could allow us to seek a  K-means-friendly latent space based clustering structure in the latent  space~\cite{yang2017towards}, it is unclear if ignoring the clustering structure information in the original space would lose useful information for the clustering membership assignment. Driven by these issues, there is a concern that a single space could lead to biased clustering membership assignments, which only consider the clustering structure based on the proxy similarity and chosen space. 

Third, even though neural networks based methods  provide the opportunity to search for a better representation, it in turn increases the  difficulty to select a suitable latent  representation space that is suitable for the task goal from different latent spaces at different epochs after the training converges\footnote{As we learn the space transformation and clustering membership assignment model in a end-to-end way, the model parameters and latent space is bound. When the representation is decided, the model is decided too.}. Unlike supervised learning that can use the ground truth to directly check the interest measurement with the validation set to help select models from different epochs, we only have training objective which is a proxy measurement, based on the clustering with proxy similarity and chosen space, in the unsupervised setting. Usually, there is no validation set,  either~\cite{xie2016unsupervised}. Our empirical results verify that there is a gap between the proxy  measurement (training  objectives) trend and the interest measurement (the task goal evaluated by  samples with ground  truth---mobility pattern types) trend in the training process. Thus, it cannot guarantee that selecting  representation space with  stop criteria based on proxy measurement could lead to desired interest measurement. So selecting any single latent space using a proxy measurement could still be biased. 

Forth, interpretability is another challenge of existing clustering algorithms. Existing algorithms usually focus on partitioning the data so that the same type of instances are expected to stay in the same cluster (and different types of instances are expected to stay in the different clusters). Most of the existing clustering algorithms provide little insight beyond clustering memberships,  limiting their interpretability and decreasing the trustworthiness of the predicted clustering membership~\cite{bertsimas2021interpretable}. 

In this paper, we propose a deep learning architecture to address the above issues. First, a permutation-equivalent operation is introduced to handle the opposite transition pattern in the same type of mobility. Second, we propose a variational autoencoder (VAE) based method that can assign clustering membership based on the clustering structure both in original space and latent space. Thus, our method can utilize an integrated similarity measure from both original and latent spaces, which is a generalization of previous similarity measures from a single space and is expected to alleviate the bias caused by chosen single space. Third, to avoid biased latent space, our final clustering membership assignment considers the clustering membership prediction from multiple latent spaces. Forth, a measurement is proposed to quantify the reliability of the predicted clustering membership by considering the dynamic behaviors of clustering membership prediction from different spaces at different epochs. This measurement could help identify the boundary areas among different clusters where the prediction is unreliable. Also, it is easier to observe incorrect predictions, improving interpretability.


\section{PRELIMINARIES}
\label{sec:preliminaries}

\subsection{Stay Points Sequence} 
\label{sec:stay_points}
\begin{figure}[!t]
\centerline{\includegraphics[width=0.76\columnwidth,height=0.12\textheight]{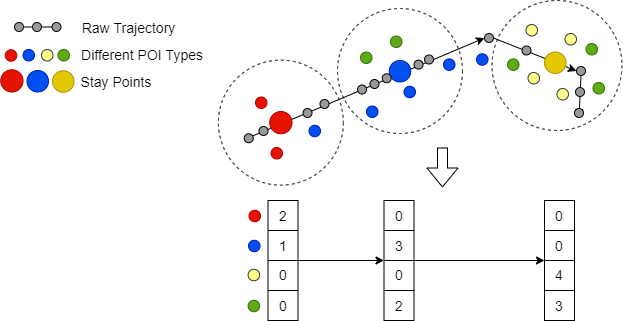}}
\caption{Stay points sequence generation~\cite{yue2021vambc}.}
\label{fig:stay_points_sequence}
\end{figure}

Using stay points sequence to preprocess raw trajectory (e.g., GPS trace) and incorporate spatial and temporal semantics has shown promising results in individual trajectory analysis~\cite{li2008mining,yue2021vambc,yue2019detect}.  Fig.~\ref{fig:stay_points_sequence} shows the procedure to generate the stay points sequence from a raw trajectory. First, the Stay Point Detection algorithm extract stay points (large colored points) from the raw trajectory  (small gray points) by partitioning the trajectory with given time duration and surrounding  area radius. Then, the Point-of-Interests (POIs) (small colored points) surrounding each stay point will be grouped by types, 
counted, and transformed into a sequence of context vectors.  Compared with the road networks based preprocessing (that aligns the GPS point into the road networks)~\cite{fang2021st2vec} or grid cells preprocessing (that partitions the whole space into equal blocks and re-maps the GPS point into the corresponding cell)~\cite{li2018deep} as the spatial context to calibrate the trajectory at geometry and geographic level, the surrounding POI information has two advantages to incorporate the spatial semantics and calibrate the trajectory at semantics level (e.g., the mobility goal): 1) it allows to easily consider similar mobility types with different transportation modalities (e.g. walking through non-road area to the school, taking subway to the school, and taking bus to the school are all school commutes) and diverse patterns (e.g. any trajectories in different schools are all school activities); 2) It is easier to attribute the individual trajectory to the human mobility intention as mobility types could be reflected from the semantics of specified areas and POI types (e.g., a stay point in school could be related to school activities behavior or school commutes) that the individual trajectories pass through. 

\subsection{Problem Definition} 

Given a set of context sequences  $\mathbf{X}=\{\mathbf{x}_i|i\in \mathbb{R}\}$, where   $x_i=\{\mathbf{v}_{i,1}, \mathbf{v}_{i,2}, ..., \mathbf{v}_{i,j}, ...,  \mathbf{v}_{i,t}\}$ is a varying-length context sequence of context vectors. The entries of the context vector  $\mathbf{v}_{i,j}$ record the frequency of the corresponding points-of-interest (POI) type in the $j$-th stay point (i.e., the POI types within the surrounding area of segmented raw trajectory). Our goal is to cluster the set of sequences $\mathbf{X}$ into K (a predefined hyper-parameter) groups/types. Mobility behavior types and trajectory types are used interchangeably for the cluster types.

\begin{figure*}[!t]
\centerline{\includegraphics[width=0.9\textwidth,height=0.16\textheight]{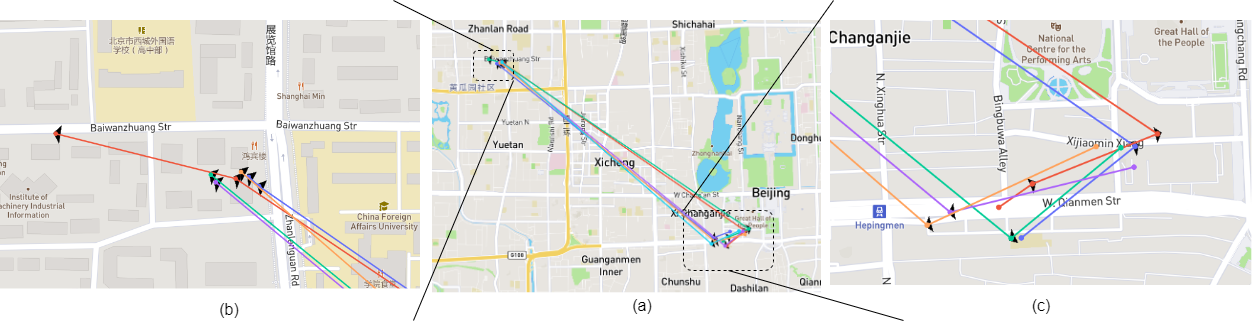}}
\caption{Opposite transition patterns in the same type of trajectories. (a) A group of trajectories with similar destination and diverse initial sub-trajectories. (b) Zoomed-in destinations. (c) Zoomed-in sub-trajectories in the beginning.}
\label{fig:opposite_pattern}
\end{figure*}

\begin{figure}[!t]
\centerline{\includegraphics[width=0.9\columnwidth,height=0.11\textheight]{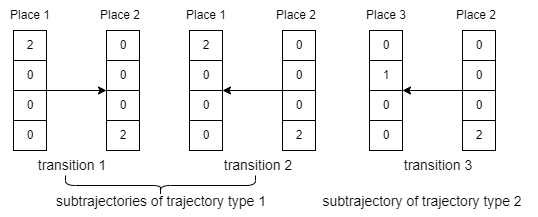}}
\caption{Transition 1 and 2 are from the same type but their start place and end place are exchanged. Transition 3 is from a different type but it shares the start with transition 2.}
\label{fig:toy_example}
\end{figure}

\subsection{Revisit Existing Trajectory Representation}  
\label{sec:trajectory_representation}

In this section, we revisit the core idea of existing methods. Even though some related methods are not limited to stay points sequence as the preprocessing step, we still include them to show the issue is widely ignored. After preprocessing, the raw trajectory is transferred into a sequence~\cite{li2018deep, yue2019detect, fang2021st2vec, yue2021vambc}. Then, two types of methods are applied. 

\noindent
\textbf{Sequence based methods:}
A trajectory is represented directly by a sequence, and the similarity is calculated by sequence matching~\cite{besse2016review,li2008mining, petitjean2011global}. 

\noindent
\textbf{Learned representation based methods:}
The trajectory is first represented by a sequence and then encoded into a vector representation. The similarity between trajectory is calculated by vector  similarity~\cite{li2018deep, yao2017trajectory, smyth1996clustering, yue2021vambc,yue2019detect,zhou2021you, fang2021st2vec}. Markov chain based method~\cite{smyth1996clustering,zhou2021you} and recurrent neural networks (RNNs)~\cite{yao2017trajectory, yue2021vambc,yue2019detect, fang2021st2vec} are the commonly used components. 

Both types of methods consider the content of elements at each time step and the temporal order explicitly by either accessing the elements to calculate the sequence similarity or encoding the sequence into a vector representation. Even though it is ``natural'' to model the temporal order, the analysis towards the effect of temporal order, especially in the trajectory clustering setting, is largely overlooked. Here, we aim to answer a question: is strictly modeling temporal transitions, retaining all the transition directions among all consecutive time steps in  the trajectory representation, to generate trajectory representation beneficial for the similarity calculation in the clustering?

To answer this question, we first revisit the widely-used existing methods. Then, we analyze the transition pattern in a real-world dataset. In addition, we discuss why existing methods fail to properly  model opposite transition patterns, which, to our best knowledge, is first observed and discussed here. Last, we propose to add a permutation-equivalent operation to help learn trajectory representations.  


\subsubsection{Opposite Transition Pattern in the Same Type of Trajectories}
We take the Geolife  dataset~\cite{zheng2010geolife} as an example for analysis. We first process the original trajectories into stay points sequences. A stay points sequence mainly consists of two parts: (1) the frequency of the POI types within the stay point at each time step; (2)  The transition between consecutive time steps. Even though both parts are important for characterizing a trajectory, we will demonstrate that strictly modeling the temporal order could decrease the trajectory representation's effectiveness in the similarity calculation.  

By mapping the stay point  sequences back to their physical positions on a map and analyzing the dataset, we  observe that \emph{some trajectories have the start places and end places of their sub-trajectories close to the end places and  start  places of other  trajectories' sub-trajectories of the same mobility behavior type,  respectively}. We call this pattern as the opposite  transition pattern in the same mobility behavior type. We visualize a group of trajectories in the   Fig.~\ref{fig:opposite_pattern}  for the example. The colors represent different trajectories. The dots  represent the stay points and the arrows represent the transition direction. Their general transitions,  starting from a small area and ending in the same block  (Fig.~\ref{fig:opposite_pattern}(a)) , indicate that they belong to the same mobility behavior.  Even though we can observe that the destinations of all these  trajectories almost locate in the same block  (Fig.~\ref{fig:opposite_pattern}(b)), these trajectories have diverse   sub-trajectories (Fig.~\ref{fig:opposite_pattern}(c)), demonstrating their opposite transition pattern for the same mobility behavior.

Now we investigate how this pattern is not handled properly by existing methods. As sequence based methods and learned representation based methods are both applied on sequences, we directly analyze the sequence for simplicity. We use a toy example without losing generality for demonstration in Fig.~\ref{fig:toy_example}.

A clustering-friendly representation should meet two clustering goals: (1) Trajectories of the same type should have similar representations; (2) Trajectories of different types should have dissimilar representations. But  strictly modeling the  temporal transition could violate both goals. Consider the example with three sub-trajectories in  Fig.~\ref{fig:toy_example}. If we consider the transition direction and sum the Euclidean distance (Other metrics or similarity could be analyzed in the similar way) at different time steps, we have $\distance(traj.1,traj.2)=2\|vec._{place1}-vec._{place2}\|_2=5.66$ and $\distance(traj.2,traj.3)=\|vec._{place1}-vec._{place3}\|_2=2.24$. This example shows that the trajectory  representations from the same type have larger distance than the  trajectory representations from different types. Thus, strictly modeling the transition will result in unexpected similarity calculation as the opposite transition pattern interferes the similarity calculation.  Also, as the stay point sequence of a trajectory is usually short (80\% of the sequence length belong in the range [2,4] in a randomly sampled GeoLife dataset), the effect of opposite transition pattern can have a major impact on the similarity calculation. Due to this ignorance, the existing generated representations hinder the clustering to achieve better results.

\subsubsection{Permutation-equivalent Operation for Trajectory Representation}

To deal with the opposite transition pattern, we  propose to add a  permutation-equivalent operation that generates equivalent  trajectory representation regardless of the effect of  transition direction. Formally,  given a context sequence, we need  to learn a mapping that  treats the input context  sequence as a set of vectors  and output a vector, where  the entries of the input  vectors belong to the  natural number range. Due to the order insensitiveness of the set, the generated  representation is  permutation-equivalent. The  most relevant recent  work~\cite{stelzner2020generative} studies learning a  permutation-equivalent mapping from a set to a vector with neural networks. Even though self-attention mechanism could be modified to retain  permutation-equivalent for a set of elements, it is not suitable to directly use their methods for a set of  vectors as self-attention only considers the  interactions among a set of  elements but a set of  vectors have interactions in both vector-level and 
entry-level. Learning a  mapping from a set of  vectors to a vector, to our best knowledge, is still an  unexplored challenging  direction. Thus, instead of directly learning a mapping,  we propose to use a fixed  operation that holds  permutation-equivalent   property and leave the learning part to the multi-layer perceptrons  after the fixed operation.  In this paper, we propose to use a \textbf{sum} operation as the permutation-equivalent mapping from the context  sequence to a vector by  directly summing the context sequence. Due to the  property of sum, the result summed vector is naturally permutation equivalent.  Still following the example in the last subsection, now we have  $\distance(traj.1,traj.2)=\distance(vec._{place1}+ vec._{place2}, vec._{place2}+ vec._{place1})=0$ and $\distance(traj.2,traj.3)=\|vec._{place1}-vec._{place3}\|_2=2.24$. Now  trajectory 1 and trajectory 2 have less distance than trajectory 2 and trajectory 3, which matches their types. 

Even though the sum operation is not a learned mapping and could limit the expression ability, we have a multi-layer perceptrons (MLP) layer just after the sum operation. This way, the proposed idea still retains strong expressive power.  And a side benefit of a sum operation is that using the sum vector as the reconstruction goal could simplify the reconstruction task. Generating a set is a challenging  task~\cite{stelzner2020generative}. And our task to generate a set of vectors is a generalization that requires the loss calculation not only to handle the permutation equivalence but also to consider the potential relations among vector-level and entry-level. The sum operation not only handles the permutation equivalent at encoding and reconstructing but also avoids the difficulty in both steps.

\section{Related Work}
\label{sec:related_work}
\noindent
\textbf{Trajectory representation and similarity:} Existing methods usually have different considerations in two folders: (1) how to effectively preprocess and incorporate spatial context into the trajectory to better achieve the application goal and (2) how to generate representations that utilize the information after prepossessing. The main goal of the prepossessing is to incorporate extra semantics to enrich or calibrate the information contained in the trajectory. Road networks based preprocessing ~\cite{fang2021st2vec}, grid cells preprocessing~\cite{li2018deep}, stay points sequence preprocessing~\cite{yue2021vambc} are commonly used methods. After preprocessing, exiting methods either use sequence based methods~\cite{besse2016review,li2008mining} or learned representation based methods~\cite{li2018deep, yao2017trajectory, smyth1996clustering, yue2021vambc,yue2019detect,zhou2021you, fang2021st2vec} to generate representation. RNN and LSTM have been the commonly-used component in the state-of-the-art methods~\cite{li2018deep, yue2021vambc,yue2019detect, fang2021st2vec}. Unlike existing methods, we first propose to use the permutation-equivalent operation to replace the RNN and LSTM due to the opposite transition pattern in the same type of human mobility. Besides, existing methods use two different ways to guide representation learning: learning representation that maximally reserves the similarity from the original space~\cite{li2018deep, fang2021st2vec} and learning algorithm-friendly representation to solve the algorithm goal (e.g., clustering-friendly representation)~\cite{dilokthanakul2016deep,yang2017towards,yue2019detect, yue2021vambc}. The similarity reservation objectives learned representation is different from the clustering goal as the original space could not be suitable for clustering and the reserved similarity objective does not consider maximizing inter-cluster distance. Our method belongs to the learning algorithm-friendly representation type and focuses on clustering. However, existing clustering-friendly representation is mainly guided by the clustering in either a single space (i.e., original space or latent space)~\cite{dilokthanakul2016deep,yang2017towards,yue2019detect} or incomplete clustering objectives~\cite{yue2021vambc}. We propose to guide the learning with clustering in both spaces to avoid the bias of a single space. Also, we propose to consider the predictions from different latent spaces after convergence to avoid the bias of a single latent space. 

As trajectory representation and similarity interact with each other in the learning process, another line of relevant work is on how existing methods deal with the similarity. Existing methods either treat the similarity as a given parameter~\cite{fang2021st2vec} or simply specify the Euclidean distance for similarity~\cite{dilokthanakul2016deep,yang2017towards,li2018deep,yue2019detect, yue2021vambc}. However, the desired definition of similarity for clustering human mobility is tricky as the task is an unsupervised analysis. It is difficult to know which distance is the best to measure human mobility at the semantics level. Thus, simply using any specified distance may constrain the analytical results due to the gap between the specified distance and desired (unknown) distance. Compared to clustering in a single space with a specified distance, clustering in both original and latent spaces in return allows us to utilize a more expressive and generalized distance that combines the specified distance in original and latent space non-trivially. As a result, we could alleviate the bias of a single space and shrink the gap by using a more expressive distance. 

\noindent
\textbf{Clustering methods:} Clustering algorithms have been studied in the past several decades~\cite{xu2015comprehensive}. Multi-view subspace  clustering~\cite{gao2015multi} assigns the cluster membership by clustering in separated independent feature subspace and aggregate the clustering results. Even though it shares a similar idea to clustering in different spaces simultaneously, they use subspaces whose union is still original space while we have multi-view from multiple spaces including original space and multiple learned latent spaces. Also, their main goal is to battle the dimension curse from large dimension while the feature space in our problem is small. Our method considers higher latent space dimension, which allows us to find a better partition for the data in the hyperspace like the kernel method for SVM. Another line of relevant research is the recent progress on the deep learning paradigm that has representation learning and clustering simultaneously. Existing methods show the superiority of this direction~\cite{xie2016unsupervised,dilokthanakul2016deep, yang2017towards}. However, their method completely ignores the clustering property on the original data space. We argue that ignoring this property can help ignore the noise in the original space, but the useful information contained in the clustering in the original space may also be sacrificed. Unlike their methods, we still use the clustering objective in the original space to help guide us find a more clustering-friendly latent representation. 

\noindent
\textbf{Clustering interpretation:} 
Existing clustering algorithms usually focus on partitioning the dataset and provide the insight of clustering membership in the dataset based on the  clustering results.  Even in this angle, most of the  existing clustering  algorithms provide little insight into the rationale for cluster membership,  limiting their  interpretability~\cite{bertsimas2021interpretable}. We propose a  metric for the reliability of the clustering membership prediction. With this metric, we can identify the boundary areas among different clusters. To overcome this limitation, a measurement is proposed to quantify the reliability of the predicted clustering membership by considering the dynamic behaviors of clustering membership prediction from
different spaces at different epochs. This measurement could
help identify the boundary areas among different clusters
where the prediction is unreliable. Also, it is easier to observe
incorrect predictions, improving interpretability.

\section{Proposed Method}
\label{sec:method}

\begin{figure}[!t]
\centerline{\includegraphics[width=1.05\columnwidth,height=0.095\textheight]{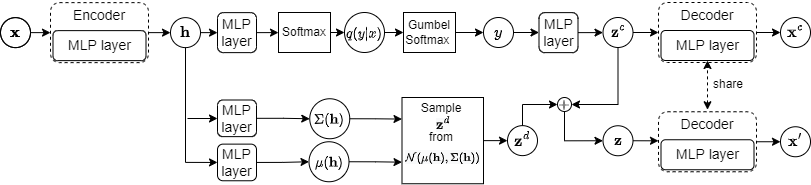}}
\caption{Network structure.}
\label{fig:network_structure}
\end{figure}

Like recent   methods~\cite{yang2017towards,yue2021vambc}, we also follow the framework that uses neural networks (in Sec.~\ref{sec:model_structure}) to learn latent spaces by considering clustering objective function and reconstruction objective function simultaneously (in Sec~\ref{sec:ojbective_function}). Our clustering membership assignment considers the clustering structure in both original space and (multiple) latent space(s). Note that the latent space in this paper is not the VAE latent  space whose goal is used for probability sampling. Our latent space is a transformed feature space (the $\mathbf{h}$ in Fig.~\ref{fig:network_structure}). To avoid misleading, we will use the VAE latent space if we talk about the probability sampling for our model. 

\subsection{Model Structure}
\label{sec:model_structure}

Fig.~\ref{fig:network_structure} shows our network structure. The input sum vector $\mathbf{x}$ is encoded into learned latent representation $\mathbf{h}$ and then $\mathbf{h}$ is transformed into VAE latent space through MLP layers. Inspired by~\cite{yue2021vambc}, the VAE latent space is decomposed into two parts: cluster VAE  latent information  $\mathbf{z}^c$ and individualized VAE latent information  $\mathbf{z}^d$.  In the top path of the network, our learned latent representation $\mathbf{h}$ is the input into a MLP layer that can predict the clustering membership $y$ with the corresponding conditional distribution $q(y|x)$ after the $\softmax$ is applied. Like~\cite{yue2021vambc}, Gumbel softmax is used to obtain the clustering membership $y$ without breaking down the back-propagation. As this path is about the cluster membership, the cluster VAE latent information $\mathbf{z}^c$ is shared by all related instances and is corresponding to the cluster center in the latent space. Thus, we only reconstruct the cluster center $\mathbf{x}^c$ from it with the decoder. In the bottom path, we have the standard VAE process which first transforms the $\mathbf{h}$ into the mean $\mu(\mathbf{h})$ and variance $\Sigma(\mathbf{h})$ of the variational variable with normal distribution $\mathcal{N}$. Then, the individualized VAE latent information vector $\mathbf{z}^d$ is sampled from this normal distribution. In order to not break down the back-propagation, the sampling step is implemented with the reparameterization trick. To avoid missing any information, the reconstruction of the input $\mathbf{x'}$ considers the combined VAE latent vector $\mathbf{z}$ from $\mathbf{z}^c$ and  $\mathbf{z}^d$.

The generative process can be described as follows: 

\begin{align}
\small
y &\backsim \Cat(\frac{1}{K}),\ \mathbf{z}^c= f(y;W),\\
\label{eq:method}
\mathbf{z}^{d}&\backsim \mathcal{N} (0,  \mathbf{I}),\ \mathbf{z} = \mathbf{z}^c + \mathbf{z}^d,\\
\mathbf{x}&\backsim \mathcal{N} (\mu(\mathbf{z}), \Sigma(\mathbf{z})),
\end{align}

The predicted clustering membership $y$ is generated by a $K$-dimensional categorical distribution where the $K$ corresponds to the number of clusters. Cluster center $\mathbf{z}^c$ in the latent space is transformed from the corresponding cluster membership $y$ and $W$ is the learnable weight matrix. The individualized VAE latent information vector $\mathbf{z}^d$ is generated from a standard normal distribution. The combined latent variable $\mathbf{z}$ is the sum of cluster center $\mathbf{z}^c$ and individualized VAE latent information vector $\mathbf{z}^d$. The input is generated from a normal distribution whose mean and variance are decided by  $\mathbf{z}$.

Compared to the network in~\cite{yue2021vambc}, the differences in our network are mainly in two major components. First, as our reconstruction target is the sum vector $\mathbf{x}$ of the stay points sequence (instead of the input sequence), our encoder and decoder are built upon MLP layers instead of RNN. Second, we use separated MLP layers to generate the mean and variance of the distribution for $\mathbf{z}^d$.

\subsection{Objective Function} 
\label{sec:ojbective_function}

Our objective function consists of two parts: reconstruction objective and clustering objective. Following the evidence lower bound in~\cite{yue2021vambc}, we have: 

\begin{align}
\small
&D_{KL}(q(y|x)||p(y)) =  -Entropy(y) + \log(K)\\
&D_{KL}(q(\mathbf{z}^d|\mathbf{x})||p(\mathbf{z}^d)) =  \| \mathbf{z}^d\|^2_2 + \V(\mathbf{z}^d) - \log\V(\mathbf{z}^d)
\label{eq:KL1}
\end{align}

\noindent
where $\V$ is the variance. So the whole reconstruction loss is:

\begin{align}
\small
\mathcal{L}_{recon} &= MSE(\mathbf{x}, \mathbf{x}') +Eq.(4) + Eq.(5)
\label{eq:recon}
\end{align}

As the original space may not be ideal for clustering, the reconstruction method is used to map the data into latent space and then conduct the clustering in the latent space~\cite{xie2016unsupervised,yang2017towards}. Existing methods~\cite{yang2017towards,yue2021vambc} of finding the latent space is not clustering-friendly enough as their objective is either only consider clustering in latent space or only consider clustering with partial clustering goal, minimizing the intra-distance. So we propose to plug in the complete clustering objective into both original space and latent space to better guide the latent space learning and form clustering assignments. 

Specifically, we have the following similar clustering objectives in original space and latent space, separately:

\begin{align}
\small
\mathcal{L}_{original} &=  \|\mathbf{x}^c - \mathbf{x}\|^2_2 - \sum\limits_{i, j\in y, i\neq j} \|\mathbf{x}_i^c -  \mathbf{x}_j^c\|^2_2,\\
\mathcal{L}_{latent} &=  \|\mathbf{z} - \mathbf{z}^c\|^2_2 - \sum\limits_{i, j\in y, i\neq j} \|\mathbf{z}_i^c -  \mathbf{z}_j^c\|^2_2
\label{eq:latent}
\end{align}

The goal of the clustering objective is to minimize the intra-cluster distance between the instance $\mathbf{x}$($\mathbf{z}$) and the corresponding center $\mathbf{x}^c$($\mathbf{z}^c$) (first term in Eq.~8 and Eq.~9) and maximize the inter-cluster distance between pairwise centers (second term in Eq.~8 and Eq.~9). Our final objective is: 

\begin{align}
\small
\argmax \mathcal{L}_{recon} +  \lambda_1\mathcal{L}_{original} + 
\lambda_2\mathcal{L}_{latent}
\label{eq:loss}
\end{align}
\\
The $\lambda_1$ and $\lambda_2$ are the hyper-parameters.  

Existing methods have shown the effectiveness of the reconstruction and clustering in the latent space to learn clustering-friendly representation~\cite{dilokthanakul2016deep, yang2017towards}. A natural question is: which kind of information is introduced by considering the clustering in the original space simultaneously? The VAE would theoretically learn the best latent space for clustering if the training process can have proper guidance (e.g., using annotated data). However, the VAE would not automatically learn useful information for clustering without proper guidance (i.e., unsupervised). The original VAE is not designed  for clustering and typically only makes “similar” data points close to each other in the latent space but cannot effectively separate “dissimilar” data points. So a clustering method is needed in the latent space. But the clustering in the latent space could be biased to the unsupervised clustering with Euclidian distance in the latent space and ignore other useful structures (e.g., important relative distance relation in the original space partially helping reflect the real similarity) in the original space that could be important for our application goal (instead of just the proxy clustering objective in the latent space). When the original dimension is small (e.g., 10 and 22 on two datasets) and the number of instances is much larger than the dimension, a higher dimensional latent space is preferred to better separate data when the data distribution is complex (e.g., the types of instances may not aggregate together and different types of instances are not separated well). But mapping data into higher dimensional latent space could potentially not retain all the relative distance relations as the distance in higher dimensional space tends to be the same (this property is the root of the dimension curse). As a result, our latent space of moderately high dimension has the potential to change the relative positions in both a good and a bad way. The good way allows data to be separated better and the bad way makes the application-goal-friendly relative distance structure in the original space broken. Thus, considering the clustering structures in both original and latent spaces, we have the separation ability from a higher dimension space to better cluster difficult instances in the latent space and retain more useful data structure information from the original space to guide learning the better latent space. As our method involves these ideas in an end-end setting, it simultaneously performs the clustering in both spaces. Considering only one clustering membership assignment exists in the method, it is easier to interpret the method as clustering with a generalized distance.  

\subsection{Inference and Ensemble Prediction} 
\label{sec:inferenc}
Although we learn clustering objectives in both original space and latent space,  we only have one clustering assignment, which can be directly calculated by $\argmax_{y} q(y|\mathbf{x})$, once the training is stopped. 

However, when a model is  learned with the proxy  measurement, it may not  guarantee performance of  the interest measurement and  makes the training stop  criteria tricky. Existing methods either stop the training when less than a certain percentage of  instances  change cluster assignment between two  consecutive epochs~\cite{xie2016unsupervised}  or overlook this step~\cite{dilokthanakul2016deep, yang2017towards,  yue2019detect, yue2021vambc}.  Our empirical analysis show that, if we continue the training after the training loss (based on proxy measurement)  converges, there is fluctuation in the  clustering  performance  (based on the interest  measurement). The  inconsistency between the proxy measurement  behavior, which is stable, and the interest  measurement, which  fluctuates, suggests that the gap  exists between training  measurement and evaluation measurement.  Similar inconsistency can be  observed  in~\cite{yue2021vambc}. If we stop training based on the proxy measurement, we may select a model that works well for the proxy  measurement but not desired for interest measurement. Thus, this  inconsistency makes the model selection tricky. We  believe this phenomenon  could also be interpreted as that the latent spaces  learned at different epochs  could be biased to the stochastic optimization on the proxy measurement and result in variance on the interest measurement. To tackle this challenge, we  propose to  ensemble  predictions from multiple  epochs after the training  converges and generate an integrated clustering assignment by considering the $q(y|\mathbf{x})$ from different epochs. This idea is inspired from the recent advance in ensemble learning with neural networks~\cite{chen2017checkpoint} which ensembles predictions from a single training process. Unlike the original goal to handle overfitting, we use ensemble  prediction from different epochs after the training converges to avoid the worst performance. Specifically, we have the ensemble prediction: 
\begin{align}
\small
\argmax_{y} \frac{\sum\limits_{i>n}q_i(y|\mathbf{x})}{max\_epoch-n+1}
\label{eq:inference}
\end{align} 
where the $n$ could be the epoch the training loss hits the local minimal or given by the user after training convergence.

\subsection{Interpretation beyond Clustering}

Existing clustering algorithms provide little insight into the rationale for cluster membership, limiting their interpretability~\cite{bertsimas2021interpretable}. Unlike the existing  focus, our clustering in multiple spaces (i.e., original space and multiple latent spaces) could provide new insights to interpret data.  Here, we introduce a new type of  interpretation---clustering membership reliability.



Besides of the side-effect of proxy measurement behavior, the training behavior can reflect data property and provide interpretation insight by  considering the different behaviors of the clustering assignment in different (latent) spaces from  different epochs. Recently, data map~\cite{swayamdipta2020dataset}  that uses the behaviors of the models on individual instances during training to yield  measurement to categorize the instances was  introduced. Their method provides the insight into the instances by  categorizing the  instances into easy-to-learn, hard-to-learn, and ambiguous types, which can help the developers to better understand the data properties in the classification  task. Inspired by their work, we first introduce a measurement into the clustering task to provide insight for the clustering membership reliability. As a result, we expect to observe the boundary area  where the different types of trajectories are close to each others.  

First, we propose a measure for the confidence level of the prediction. We define the \textbf{confidence} of the clustering membership for instance $i$ as the mean  probability of the predicted membership $y^*$ with  formula~\ref{eq:inference} as: 
$$\hat{\mu}_i = \frac{1}{max\_epoch-n+1}\sum\limits_{e=n}^{max\_epoch}q_e(y_i^*|\mathbf{x}_i),$$  where the $n$ could be the epoch the training loss hits the local minimal or some parameter given by the user. We only consider partial epochs at the end of the training process, which is different from classification tasks that consider all the training epochs~\cite{swayamdipta2020dataset}. We do not consider early epochs during training because the latent spaces are not learned well during these steps. These early latent spaces introduce more noise than useful information.  Intuitively, a high-confidence instance is ``easier'' for prediction. 

We define \textbf{variability}  for the changes in the prediction probability among different (latent) spaces by  standard deviation:
$$\hat{\sigma}_i = \sqrt {\frac{\sum\limits_{e=n}^{max\_epoch}(q_e(y_i^*|\mathbf{x}_i)-\hat{\mu}_i)^2}{max\_epoch-n+1}}.$$ 
An instance assigned by with the same clustering membership consistently (whether accurately or not) has low variability. 

Finally, we define \textbf{reliability} as the ratio: $r_i=\frac{\hat{\mu}_i}{\hat{\sigma}_i}.$
Intuitively, high reliability implies the instance is predicted confidently with the clustering membership. We will show that by comparing the reliability of all predictions, we could identify the boundary area among different clusters, where it is easy to observe that different types of trajectories are close to each others. And most of the incorrect predictions can be observed in/near the boundary area.

\section{Experimental Results}
\label{sec:experiment}

\subsection{Experiment  Setting}
\textbf{Dataset:} Following~\cite{yue2021vambc,yue2019detect},  we use the GeoLife dataset~\cite{zheng2010geolife} and DMCL dataset\footnote{\url{https://www.cs.uic.
edu/~boxu/mp2p/gps_data.html}}, which contain the GPS data of real human trajectories. Unlike taxi trajectories analysis work  ~\cite{li2018deep,  fang2021st2vec}, we focus on individual trajectories that are more general to reflect human mobility. The POI information of two datasets are from  OpenStreetMap  (OSM)\footnote{\url{http://download.geofabrik.de/north-america.html}} 
and the PKU Open Research Data\footnote{\url{https://doi.org/10.18170/DVN/WSXCNM}
},  respectively. After preprocessing, the staypoint sequence lengths are in the range [2, 9] and [2, 6], respectively. More than 86\% of sequences are in the [2, 4] length. There are 22 and 10 POI  types, respectively. 
\\
\textbf{Evaluation metrics (the interest measurement):} Considering the application goal to check the effectiveness of the methods to group ``similar'' individual mobility and the desired similarity is difficult to define and choose, we use the mobility types as the groundtruth labels and compare them with our predicted membership for evaluation. As the original datasets do not have the labels, the experiments are evaluated  based on the 600 and 100  labeled trajectories samples from the GeoLife and DMCL datasets,  respectively~\cite{yue2019detect}.  According to the human urban mobility activities analysis~\cite{jiang2012clustering}, six labels were provided for GeoLife dataset as the groundtruth classes: ``campus activities'', ``hangouts'', ``dining activities'', ``healthcare
activities'', ``working commutes'', ``school commutes''.  Four labels were provided as the groundtruth classes for DMCL dataset: ``school commutes'', ``residential activities'', ``campus
activities'', ``hangouts''. Two widely-used clustering metrics are applied for evaluation: Normalized  Mutual Information (NMI) and  Clustering Accuracy (Acc)   ~\cite{xie2016unsupervised, dilokthanakul2016deep, yang2017towards,  yue2019detect,  yue2021vambc}.  
\\
\textbf{Compared methods:} All the compared methods are evaluated on the stay points sequence after preprocessing. Some recent methods learning similarity reservation trajectory representation ~\cite{li2018deep,fang2021st2vec} are not considered as their methods tightly rely on different prepossessing to calibrate the trajectory. Also, their representations focus on approximating the similarity in the original space, while we focus on clustering problem in which maximizing inter-cluster distance is also critical and could modify the relative similarity in the original space. \\
\textbf{(1) Sequence based methods:}
traditional methods.\\
\textbf{\emph{MHMM}}~\cite{smyth1996clustering}: It is a generative model that can generate a sequence of multivariate feature vectors. The model is based on the mixture model and the hidden Markov model. The clustering is applied to the log-likelihood distance matrix.
\\
\textbf{\emph{KM-DTW}}~\cite{petitjean2011global}: It uses DTW to calculate the distance between sequences and then uses K-means clustering. 
\\
\textbf{(2) Learned representation based methods:} Several state-of-the-art deep learning-based clustering methods or trajectory clustering methods are tweaked or used for comparison. 
\\    
\textbf{\emph{RNN+GMVAE}}~\cite{dilokthanakul2016deep}: It is a VAE based clustering model with a Gaussian mixture as a prior distribution. RNN is used to handle the sequence and reconstruct the sequence.  
\\
\textbf{\emph{RNN+DCN}}~\cite{yang2017towards}: It is based on auto-encoder and K-means. The method learns latent representation by reconstructing the feature vector and clustering with neural network based K-means in the latent space,  simultaneously. A sophisticated optimizer is proposed to learning their network. RNN is used to handle the sequence and reconstruct the sequence.
\\
\textbf{\emph{DETECT}}~\cite{yue2019detect}: It learns latent representation by reconstructing the feature vector and clustering with neural network based K-means in the latent space. The reconstructing and clustering perform separately in two phrases. Unlike DCN, auxiliary target to minimize the clustering cleanness is used. RNN is used to handle the sequence and reconstruct the sequence.
\\
\textbf{\emph{VAMBC}}~\cite{yue2021vambc}: It is the state-of-the-art trajectories clustering method based on VAE and neural networks based clustering. It learns latent  representation by reconstructing the sequence and a more general neural networks based clustering. Unlike DETECT, the reconstructing and clustering are performed simultaneously. It uses RNN as encoder and decoder.   
\\
\textbf{(3) SUM operation based methods:} To verify the effectiveness of SUM operator, we apply it with state-of-the-art deep learning based clustering methods mentioned above.
\\
\textbf{\emph{SUM+K-means}}: It applies K-means on the sum vector. This directly shows the effect of the permutation-equivalent without being affected by a sophisticated clustering algorithm.\\
\textbf{\emph{SUM+GMVAE}}: The GMVAE is applied on the sum vector. \\
\textbf{\emph{SUM+DCN}}: the DCN is applied on the sum vector.  

The number of clusters is determined by the elbow method. It detects the correct cluster numbers in both datasets. For the learning-based methods, we tune the parameters (e.g., the layers of neural networks, the size of the latent dimension, the optimizer and corresponding parameters, and so on). If the compared methods do not provide stop criteria and we observe better performance on interest measurement after training convergence, we give additional enough epochs and select the epoch with the best interest measurement. Even though this is not practical in the real application (unsupervised setting), we argue that it is not obvious to find a proper stop condition due to the gap between training loss and the interest measurement. This can avoid the performance decrease caused by the stop condition. The parameter tuning for the proposed method is based on the training loss. Additional 100--200 epochs after training convergence are given to the proposed method, and the ensemble prediction is based on these extra epochs.

\subsection{Clustering Performance}

\begin{table}[ht]
\small
\caption{The comparison on Geolife and DMCL datasets.}
\resizebox{0.5\textwidth}{!}{
\begin{tabular}{l|cc|cc|cc|cc}
\toprule 
Data&Geolife &Acc. & &NMI &DMCL &Acc. & &NMI\\
  &Ave.& Max &Ave.& Max &Ave.& Max &Ave.& Max\\ 
 \hline 
  SUM+K-means  &0.7958	&0.8015

 &0.6857&0.6982 &0.6082	&0.6121

 &0.4279&0.4686
 \\

 KM-DTW & 0.7426& 0.7638&0.6104&0.6451 &0.5827&0.6004&0.3662&0.4155
 \\

 MHMM &0.6275& 0.6497&0.5305& 0.6113 &0.6483&0.7565&0.3264& 0.3926
 \\

RNN+GMVAE &0.5307&0.6173&0.4475&0.5988& 0.5662&0.6227&0.3195& 0.4769
\\
SUM+GMVAE &0.5845&0.6589&0.5075&0.6419&0.5972&0.6657&0.4636&0.4451
\\

RNN+DCN 
&0.7826&0.8408&0.6462&0.7258&0.7853&\textbf{\underline{0.8037}}&0.4476&0.4791
\\
SUM+DCN 
&0.8026&\textbf{\underline{0.8436}}&0.6964&\textbf{\underline{0.7312}}&0.7907&0.8014&0.5041&\textbf{\underline{0.5425}}
\\

DETECT 
&0.8002 &0.8220 &0.6445 &0.6912&0.7791 &0.8002 &0.4864 &0.5270
\\

VAMBC
&\textbf{\underline{0.8251}}	&0.8425
 &\textbf{\underline{0.6973}}&0.6992
&\textbf{\underline{ 0.7993}}	&0.8005
 &\textbf{\underline{0.5129}}&0.5275
 \\

 Proposed  &\textbf{0.8469}	&\textbf{0.8768}	

 &\textbf{0.7512}&\textbf{0.7812}&\textbf{0.8234}	&\textbf{0.8368}	

 &\textbf{0.6206}&\textbf{0.6427}

\\

\bottomrule
\end{tabular}
}
\label{exp:comparison_geolife}
\end{table}




 









Tables~\ref{exp:comparison_geolife} shows the experiment results. The value with \textbf{bold} is the highest in each column. The value with  \textbf{\underline{underline}} is the second highest.  Several observations can be made. First, the proposed method  consistently outperforms other methods in both datasets. It indicates the effectiveness of our proposed method. Second, the variants with the sum operation achieve competitive performance (sometimes better), compared to the corresponding methods with RNNs. It implies that the proposed sum operation brings considerable  benefit. Third, the  clustering performance in the original space  and latent space varies a lot  across different datasets. The simple baseline SUM+K-means achieves very competitive performance close to the state-of-the-art method VAMBC in the Geolife. However, the performance decreases in DMCL dataset as the DMCL dataset has more noises  (e.g., outlines)  and K-means cannot handle noisy data well. However, when clustering is performed in the latent space by SUM+DCN, the performance stay high across different datasets.  It implies that the  clustering structure in the original space could be useful but it may be  impacted by other issue (e.g. noise in the original space). And our method can  utilize the useful information. Forth, the learned representation based methods can achieve much better performance than the sequence based methods.  


\begin{figure}[!t]
\centerline{\includegraphics[width=1.05\columnwidth,height=0.24\textheight]{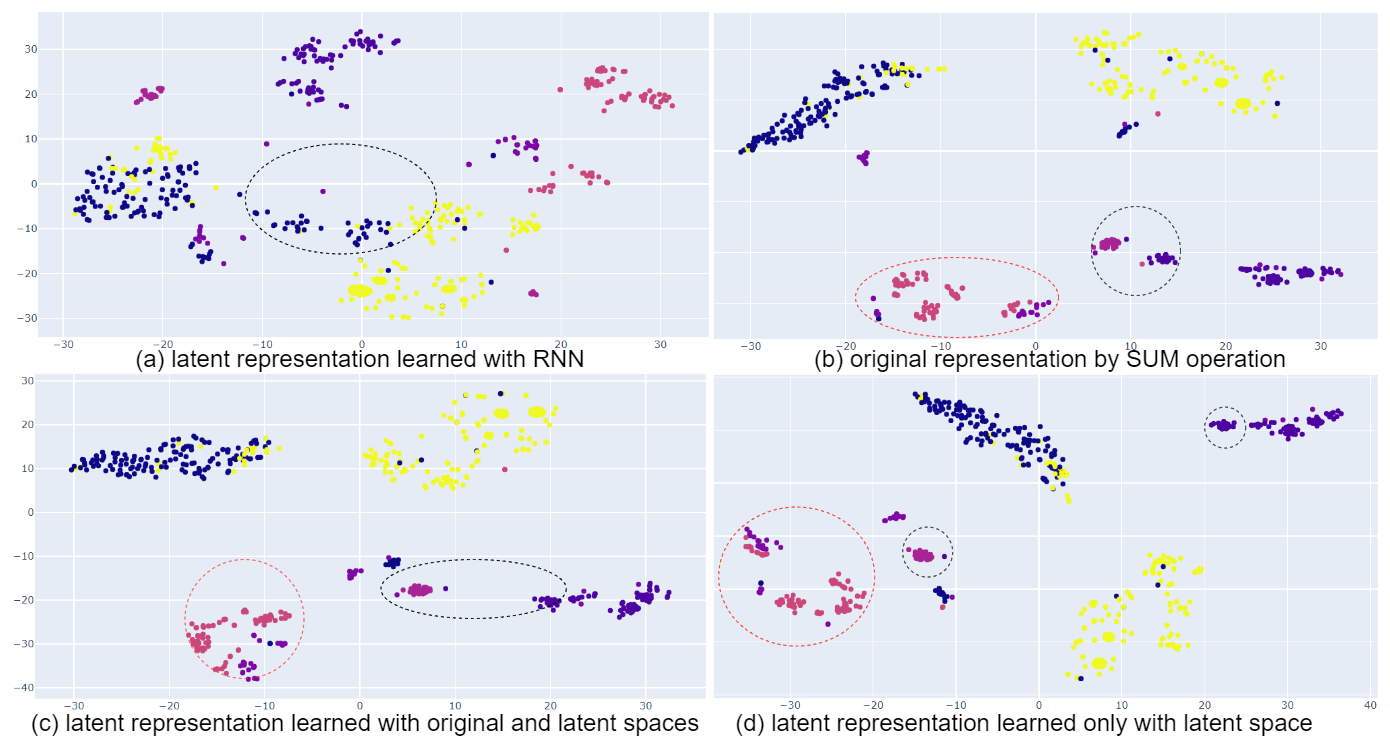}}
\caption{The t-SNE plots for the  trajectory representation in the different spaces. Colors are the ground truth types.}
\label{fig:original_effect}
\end{figure}

\begin{figure}[!t]
\centerline{\includegraphics[width=0.95\columnwidth,height=0.12\textheight]{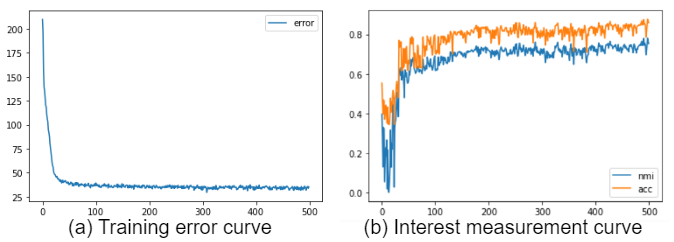}}
\caption{Training process. The x axis is epoch and y axis is error/metric.}
\label{fig:training_process}
\end{figure}

\begin{figure}[!t]
\centerline{\includegraphics[width=1\columnwidth,height=0.12\textheight]{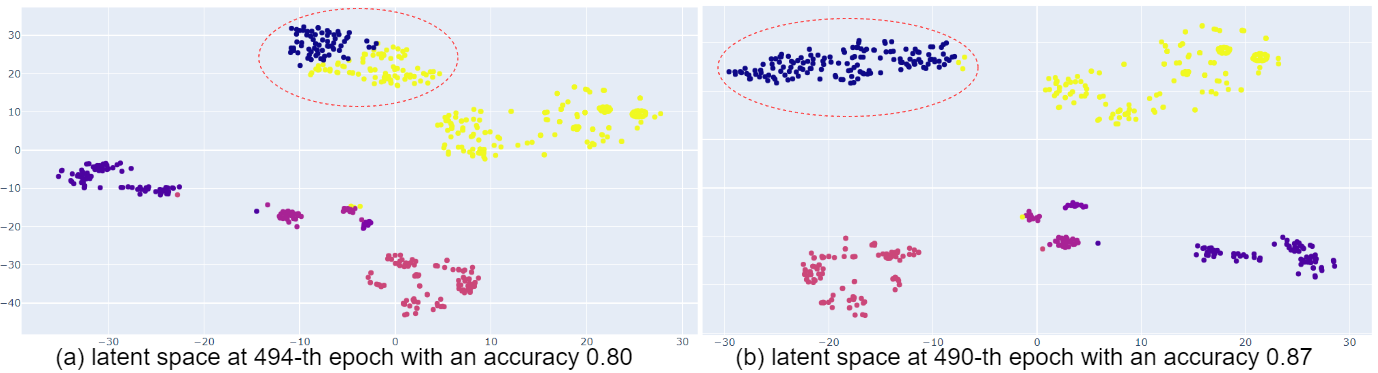}}
\caption{The t-SNE plots for latent spaces at different epochs. Colors are the predicted clustering memberships.}
\label{fig:different_epochs}
\end{figure}

\subsection{Ablation Study}
\noindent
Only the study on GeoLife is shown due to space limitation. 
\\
\textbf{The effect of the sum operation:} The quantity results of the variants with SUM operator as the prefix in the last section have shown the effectiveness. For additional quality analysis, we visualize and compare the quality of the representations generated by RNN and sum operation. To focus on RNN and sum, we remove the clustering objective and the proposed network structure. We simply use an auto-encoder with the RNN as the encoder and decoder to reconstruct the sequence to generate the latent representation and compare it with the sum vector representation. The RNN representation is shown in the Fig.~\ref{fig:original_effect}(a)  and the sum representation is shown in Fig.~\ref{fig:original_effect}(b) with the t-SNE plot. We can observe that the blue instances in the RNN representation space are more discrete (especially in the black circle of Fig.~\ref{fig:original_effect}(a)) than the counterparts in the sum vector representation space. We did observe that many opposite transition patterns exist in the blue trajectories. It verifies the effectiveness of the sum operation.
\\
\textbf{The effect of the clustering structure in original space:} 
We analyze how the  clustering structure in the original space affects the learned latent space. We compare the proposed method with a variant that is trained without the original space. We show the results from a randomly selected epoch (after convergence) for both methods. The proposed method learned with both original space and latent space achieves accuracy 0.85, while the variant only achieves 0.73. This gap supports the effectiveness of considering the clustering structure in the original spaces. Careful readers may realize that, even though SUM+DCN (accuracy 0.80) shares a similar idea with our variant, their performances are different. Actually, their implementations are different. First, DCN has its own carefully designed optimizer while we only use the commonly-used Adam optimizer. Second, DCN does not provide the training stop condition, but we observe the clustering accuracy still goes up after training convergence. Thus, we give enough epochs and report the best accuracy DCN achieves. But for our variant, we sample a space that could not be the best and report the corresponding accuracy. Even compared with the method with more optimizations, clustering in both spaces consistently improves the clustering performance.

To better understand, we visualize and compare the latent spaces in these two ways. Even though our method use ensemble prediction, we observe that the latent spaces after training convergence with different performance have similar data distribution in general (Fig.~\ref{fig:different_epochs}).  So we show the randomly selected latent spaces as representative examples in Fig.~\ref{fig:original_effect}. We can make several observations,  by comparing  Fig.~\ref{fig:original_effect}(b)(c)(d).  First,  it is obvious that the purple instances (in the red circle) aggregate in a group when we consider both spaces (in the Fig.~\ref{fig:original_effect}(c)) while they are separated when we only consider either the original space  (Fig.~\ref{fig:original_effect}(b)) or the latent space Fig.~\ref{fig:original_effect}(d). Second, the relative position  of the clusters in the original space are generally retained in the latent space when we consider both spaces. But the relative positions of some clusters change when we only consider the clustering in the latent space. For example, if we consider the two clusters in the black circles, they are close to each other in the original space (in Fig.~\ref{fig:original_effect}(b)). And this structure is generally retained in the Fig.~\ref{fig:original_effect}(c). Their distance is enlarged a little as they belong to different clusters. But compared to the  distances to other  clusters, the distance between the clusters in the black circle is not large. However, this structure is  destroyed in the Fig.~\ref{fig:original_effect}(d). It implies that when we consider both clustering structures, the learned latent space can retain more semantics from the original space, which could find better latent space.
\\
\textbf{The training dynamics and different latent spaces:} We analyze the training process and show the effectiveness of considering multiple latent spaces. Note that, unlike the ensemble prediction in classification to achieve better generalization on the test set, the goal of the ensemble prediction in our clustering task is to avoid selecting only a biased (and not good enough) latent space and the corresponding undesired clustering performance. Our training curves are shown in Fig.~\ref{fig:training_process}. We can observe that the training error converges quickly and stably in Fig.~\ref{fig:training_process}(a). However, the metrics of interest fluctuate after the training converges (around 300th epoch) in Fig.~\ref{fig:training_process}(b). Due to this fluctuation, we may select a model with undesired performance if we cannot have a proper training stop condition. This supports the importance of our ensemble prediction by averaging the epochs after training convergence. In Fig.~\ref{fig:different_epochs}, we visualize two latent spaces at different epochs for additional analysis. Even though the latent spaces from different epochs have similar data distribution in general, some details are different in the red circles. The clusters in Fig.~\ref{fig:different_epochs}(a) seem to be more compact, and the distance between instances seems to be smaller distances than the corresponding elements in Fig.~\ref{fig:different_epochs}(b). Thus, the latent space in (a) is biased to produce instances with short distances and predict more yellow in the red circle compared to the latent space in (b). Selecting any single latent space could result in a biased space that may provide undesired performance. To avoid this, our method considers different latent spaces by aggregating the clustering membership prediction from multiple latent spaces.

\subsection{Interpretation of Reliability}

\begin{figure}[!t]
\centerline{\includegraphics[width=1\columnwidth,height=0.19\textheight]{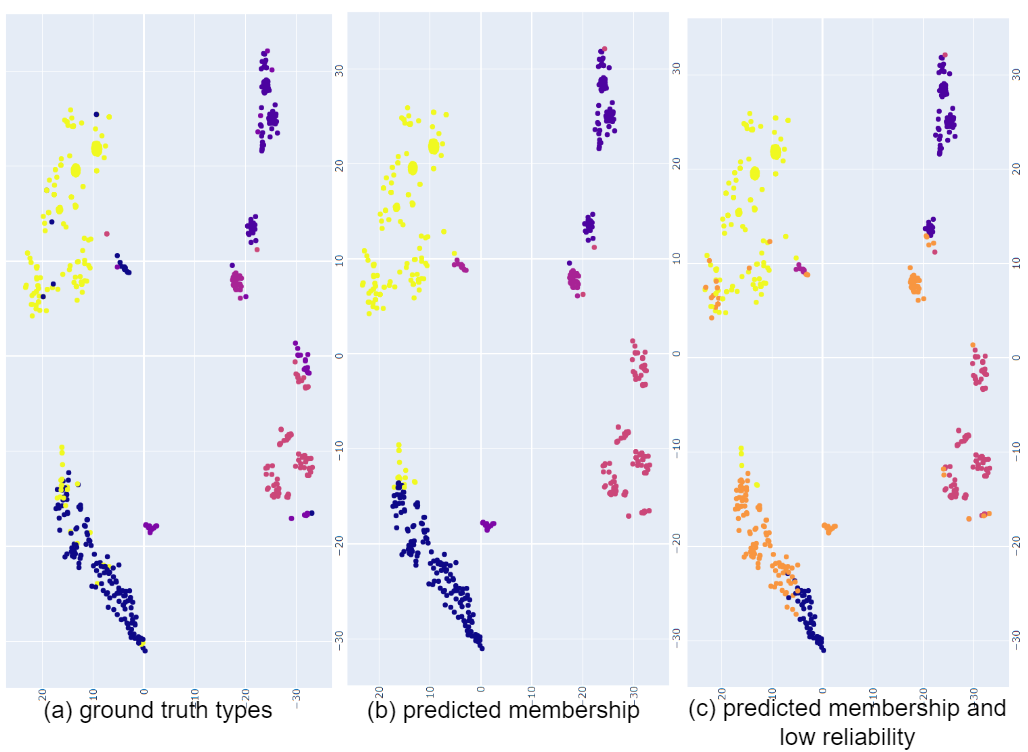}}
\caption{The t-SNE plots for the trajectory representation in the original space: (a) colors are the ground truth types; (b) Colors are the predicted clustering memberships; (c) Same as (b), except that the orange instances mean their reliability is below a threshold. The threshold is decided by all the reliability values.}
\label{fig:reliability}
\end{figure}

To show the insights of reliability, we plot the t-SNE for all the trajectory representations with  different colors representing (a) the ground truth trajectory types, (b) the predicted clustering memberships, and (c) the low reliability in the  Fig.~\ref{fig:reliability}. We can observe that the low reliability instances can reflect  the boundary areas among different trajectory types. The instances with different colors are close to each others within the orange areas.  Also, most of the incorrect membership assignments can be observed  there (by comparing Fig.~\ref{fig:reliability}(a) and (b)). There are a small number of incorrectly predicted instances not covered, but they are close enough to the boundary areas.

\section{Conclusion}
\label{sec:conclusion}

In this paper, we propose a novel solution for clustering human mobility. First, the proposed SUM operator not only helps generate permutation-equivalent representation but also simplifies reconstruction. Second, the proposed reconstruction and clustering in both original and latent spaces objectives could guide a better representation learning and generate more accurate clustering prediction. Third, due to the gap between the training objective and interest measurement, an ensemble method upon a single training process is proposed to help avoid undesired performance caused by improper (and hard-to-design) training stop conditions. Forth, a metric based on the prediction behaviors from different latent spaces at different epochs after training convergence is proposed as the prediction reliability and helps detect the boundary area where different types of trajectories are close to each other. The experiments verify the effectiveness of the proposed method. 

\bibliographystyle{IEEEtranS}
\bibliography{icdm19.bib}

\begin{thebibliography}{10}
\providecommand{\url}[1]{#1}
\csname url@samestyle\endcsname
\providecommand{\newblock}{\relax}
\providecommand{\bibinfo}[2]{#2}
\providecommand{\BIBentrySTDinterwordspacing}{\spaceskip=0pt\relax}
\providecommand{\BIBentryALTinterwordstretchfactor}{4}
\providecommand{\BIBentryALTinterwordspacing}{\spaceskip=\fontdimen2\font plus
\BIBentryALTinterwordstretchfactor\fontdimen3\font minus
  \fontdimen4\font\relax}
\providecommand{\BIBforeignlanguage}[2]{{%
\expandafter\ifx\csname l@#1\endcsname\relax
\typeout{** WARNING: IEEEtranS.bst: No hyphenation pattern has been}%
\typeout{** loaded for the language `#1'. Using the pattern for}%
\typeout{** the default language instead.}%
\else
\language=\csname l@#1\endcsname
\fi
#2}}
\providecommand{\BIBdecl}{\relax}
\BIBdecl

\bibitem{bertsimas2021interpretable}
D.~Bertsimas, A.~Orfanoudaki, and H.~Wiberg, ``Interpretable clustering: an
  optimization approach,'' \emph{Machine Learning}, vol. 110, no.~1, 2021.

\bibitem{besse2016review}
P.~C. Besse, B.~Guillouet, J.-M. Loubes, and F.~Royer, ``Review and perspective
  for distance-based clustering of vehicle trajectories,'' \emph{IEEE Trans. on
  Intelligent Transportation Systems}, vol.~17, no.~11, 2016.

\bibitem{chen2017checkpoint}
H.~Chen, S.~Lundberg, and S.-I. Lee, ``Checkpoint ensembles: Ensemble methods
  from a single training process,'' \emph{AAAI}, 2018.

\bibitem{dilokthanakul2016deep}
N.~Dilokthanakul, P.~A. Mediano, M.~Garnelo, M.~C. Lee, H.~Salimbeni,
  K.~Arulkumaran, and M.~Shanahan, ``Deep unsupervised clustering with gaussian
  mixture variational autoencoders,'' \emph{arXiv preprint arXiv:1611.02648},
  2016.

\bibitem{fang2021st2vec}
Z.~Fang, Y.~Du, X.~Zhu, L.~Chen, Y.~Gao, and C.~S. Jensen, ``St2vec:
  Spatio-temporal trajectory similarity learning in road networks,''
  \emph{arXiv preprint arXiv:2112.09339}, 2021.

\bibitem{gao2015multi}
H.~Gao, F.~Nie, X.~Li, and H.~Huang, ``Multi-view subspace clustering,'' in
  \emph{Proc. of the IEEE international conference on computer vision}, 2015.

\bibitem{gonzalez2008understanding}
M.~C. Gonzalez, C.~A. Hidalgo, and A.-L. Barabasi, ``Understanding individual
  human mobility patterns,'' \emph{nature}, vol. 453, no. 7196, 2008.

\bibitem{jiang2012clustering}
S.~Jiang, J.~Ferreira, and M.~C. Gonz{\'a}lez, ``Clustering daily patterns of
  human activities in the city,'' \emph{Data Mining and Knowledge Discovery},
  vol.~25, no.~3, 2012.

\bibitem{li2008mining}
Q.~Li, Y.~Zheng, X.~Xie, Y.~Chen, W.~Liu, and W.-Y. Ma, ``Mining user
  similarity based on location history,'' in \emph{Proceedings of the 16th ACM
  SIGSPATIAL international conference on Advances in geographic information
  systems}, 2008.

\bibitem{li2018deep}
X.~Li, K.~Zhao, G.~Cong, C.~S. Jensen, and W.~Wei, ``Deep representation
  learning for trajectory similarity computation,'' in \emph{2018 IEEE 34th
  international conference on data engineering (ICDE)}, 2018.

\bibitem{petitjean2011global}
F.~Petitjean, A.~Ketterlin, and P.~Gan{\c{c}}arski, ``A global averaging method
  for dynamic time warping, with applications to clustering,'' \emph{Pattern
  recognition}, vol.~44, no.~3, 2011.

\bibitem{smyth1996clustering}
P.~Smyth, ``Clustering sequences with hidden markov models,'' \emph{Advances in
  neural information processing systems}, vol.~9, 1996.

\bibitem{stelzner2020generative}
K.~Stelzner, K.~Kersting, and A.~R. Kosiorek, ``Generative adversarial set
  transformers,'' in \emph{Workshop on Object-Oriented Learning at ICML},
  vol.~3, 2020.

\bibitem{su2021unveiling}
R.~Su, E.~C. McBride, and K.~G. Goulias, ``Unveiling daily activity pattern
  differences between telecommuters and commuters using human mobility motifs
  and sequence analysis,'' \emph{Transportation Research Part A: Policy and
  Practice}, vol. 147, 2021.

\bibitem{swayamdipta2020dataset}
S.~Swayamdipta, R.~Schwartz, N.~Lourie, Y.~Wang, H.~Hajishirzi, N.~A. Smith,
  and Y.~Choi, ``Dataset cartography: Mapping and diagnosing datasets with
  training dynamics,'' \emph{EMNLP}, 2020.

\bibitem{xie2016unsupervised}
J.~Xie, R.~Girshick, and A.~Farhadi, ``Unsupervised deep embedding for
  clustering analysis,'' in \emph{International conference on machine
  learning}.\hskip 1em plus 0.5em minus 0.4em\relax PMLR, 2016.

\bibitem{xu2015comprehensive}
D.~Xu and Y.~Tian, ``A comprehensive survey of clustering algorithms,''
  \emph{Annals of Data Science}, vol.~2, no.~2, 2015.

\bibitem{yang2017towards}
B.~Yang, X.~Fu, N.~D. Sidiropoulos, and M.~Hong, ``Towards k-means-friendly
  spaces: Simultaneous deep learning and clustering,'' in \emph{international
  conference on machine learning}.\hskip 1em plus 0.5em minus 0.4em\relax PMLR,
  2017.

\bibitem{yao2018learning}
D.~Yao, C.~Zhang, Z.~Zhu, Q.~Hu, Z.~Wang, J.~Huang, and J.~Bi, ``Learning deep
  representation for trajectory clustering,'' \emph{Expert Systems}, vol.~35,
  no.~2, 2018.

\bibitem{yao2017trajectory}
D.~Yao, C.~Zhang, Z.~Zhu, J.~Huang, and J.~Bi, ``Trajectory clustering via deep
  representation learning,'' in \emph{2017 international joint conference on
  neural networks (IJCNN)}, 2017.

\bibitem{yue2021vambc}
M.~Yue, Y.-Y. Chiang, and C.~Shahabi, ``Vambc: A variational approach for
  mobility behavior clustering,'' in \emph{Joint European Conference on Machine
  Learning and Knowledge Discovery in Databases}, 2021.

\bibitem{yue2019detect}
M.~Yue, Y.~Li, H.~Yang, R.~Ahuja, Y.-Y. Chiang, and C.~Shahabi, ``Detect: Deep
  trajectory clustering for mobility-behavior analysis,'' in \emph{2019 IEEE
  International Conference on Big Data (Big Data)}, 2019.

\bibitem{zheng2010geolife}
Y.~Zheng, X.~Xie, W.-Y. Ma \emph{et~al.}, ``Geolife: A collaborative social
  networking service among user, location and trajectory.'' \emph{IEEE Data
  Eng. Bull.}, vol.~33, no.~2, 2010.

\bibitem{zhou2021you}
Y.~Zhou, Q.~Yuan, C.~Yang, and Y.~Wang, ``Who you are determines how you
  travel: Clustering human activity patterns with a markov-chain-based mixture
  model,'' \emph{Travel Behaviour and Society}, vol.~24, 2021.

\end{thebibliography}


\end{document}